%% file: acl_latex.tex
\pdfoutput=1

\documentclass[11pt]{article}

\usepackage[final]{acl}

\usepackage{times}
\usepackage{latexsym}

\usepackage[T1]{fontenc}

\usepackage[utf8]{inputenc}

\usepackage{microtype}

\usepackage{inconsolata}

\usepackage{graphicx}

\usepackage{hyperref}       
\usepackage{url}            
\usepackage{booktabs}       
\usepackage{amsfonts}       
\usepackage{nicefrac}       
\usepackage{xcolor}         

\usepackage{makecell}
\usepackage{tabu}
\usepackage{tabularx}
\usepackage{caption}
\usepackage{subfigure}
\usepackage{multirow}
\usepackage{amsmath}

\title{LLM Reasoning Engine: Specialized Training for \\ Enhanced Mathematical Reasoning}

\author{%
    Shuguang Chen\\
    Department of Mathematics\\
    Purdue University\\\
    \texttt{chen4914@purdue.edu} \\
    \And
    Guang Lin \\
    Department of Mathematics\\
    Purdue University\\
    \texttt{guanglin@purdue.edu} \\
}

\begin{document}
\maketitle
\begin{abstract}
Large Language Models (LLMs) have shown remarkable performance in various natural language processing tasks but face challenges in mathematical reasoning, where complex problem-solving requires both linguistic understanding and mathematical reasoning skills. Existing approaches to address this challenge often rely on ensemble methods and suffer from the problem of data scarcity in target domains. In this work, we present a novel method to enhance the capabilities of LLMs in mathematical reasoning tasks. Motivated by the need to bridge this gap, our approach incorporates a question paraphrase strategy, which aims to diversify the linguistic forms of mathematical questions to improve generalization. Additionally, specialized training objectives are employed to guide the model's learning process, focusing on enhancing its understanding of mathematical concepts and reasoning processes. We conduct experiments on four datasets using different LLMs, and demonstrate the effectiveness of our approach in improving LLMs' performance on mathematical reasoning tasks. Our findings underscore the significance of our methodology in advancing LLMs and their potential implications for real-world applications that require mathematical reasoning abilities.
\end{abstract}

\section{Introduction}

In recent years, Large Language Models (LLMs) \citep{touvron2023llama2, team2023gemini} have emerged as powerful tools in the field of machine learning, demonstrating remarkable performance in a wide range of downstream benchmarks. Their ability to understand and generate natural language text has revolutionized various applications, from language translation \citep{costa2022no, barrault2023seamlessm4t} to question answering systems \citep{chowdhery2023palm, chen2023meditron}. Central to their success is their ability to perform complex reasoning, enabling them to tackle complex problems with impressive accuracy and efficiency.

However, while LLMs excel in many domains, they face challenges when it comes to solving mathematical problems. Mathematical reasoning often requires intricate logical operations and a deep understanding of mathematical concepts \citep{saxton2019analysing, lightman2023let}, posing significant hurdles for conventional LLM architectures. Moreover, the scarcity of data in the mathematical domain \citep{liu2021roda, kumar2022practice} further compounds these challenges, limiting the performance and generalization of these LLMs.

\input{figures/illustration_example}

Some existing methods \citep{shen2023mixture, jiang2024mixtral} attempt to address these challenges by employing ensemble techniques, where multiple LLMs collaboratively solve mathematical problems. However, these methods can still be further improved, particularly in enhancing the performance of individual LLMs and mitigating performance degradation over extended reasoning steps. A major challenge when applying LLMs to complex mathematical problems is error propagation, especially in tasks requiring long reasoning chains. Figure \ref{fig: illustration_example} illustrates this issue with an example from the GSM8K dataset \citep{cobbe2021gsm8k}, solved by the Llama2-7B model \citep{touvron2023llama2}. The problem involves multiple intermediate steps to reach the correct solution. However, the model makes an error in the initial steps, which cascades through subsequent stages of reasoning. This occurs because each reasoning step depends on the accuracy of preceding steps; even a minor initial error can compound, ultimately leading to significant deviations from the correct solution. This phenomenon highlights the limitations of current LLMs in maintaining accuracy across long reasoning sequences and underscores the need for improved methodologies to mitigate error propagation and enhance performance in mathematical tasks.

In this paper, we propose novel approaches to address the limitations of existing methods in solving mathematical problems with LLMs. Our contributions include the introduction of new training objectives designed to uncover underlying patterns in data, thus improving model performance. Additionally, we leverage data augmentation techniques to maximize the utility of existing datasets to fine-tune LLMs, enhancing their effectiveness in mathematical reasoning tasks. To evaluate the efficacy of our proposed methods, we conduct experiments using four open-source LLMs—Llama \citep{touvron2023llama}, Llama2 \citep{touvron2023llama2}, Mistral \citep{jiang2023mistral}, and Mixtral \citep{jiang2024mixtral}—on four widely used mathematical reasoning datasets: GSM8K \citep{cobbe2021gsm8k}, MATH \citep{hendrycksmath2021}, GSM8K\_Hard \citep{gao2022pal}, and SVAMP \citep{patel-etal-2021-nlp}. Our results demonstrate significant performance improvements, underscoring the effectiveness of our methods in advancing the mathematical reasoning capabilities of LLMs.

In summary, this paper presents novel contributions towards improving the performance of LLMs in solving mathematical problems, addressing the challenges posed by complex reasoning and data scarcity. Our findings have implications for advancing the field of machine learning and expanding the applicability of LLMs to various problem domains.

\section{Background}
The intersection of machine learning and mathematical reasoning has received significant attention from researchers seeking to enhance the capabilities of LLMs in solving complex mathematical problems. In this section, we review the previous literature exploring various approaches and methodologies to address the challenges posed by mathematical reasoning tasks and highlight the advances made in this domain.

\subsection{Recent Advances in LLMs}
The evolution of LLMs represents a significant milestone in the field of natural language processing (NLP) and machine learning. Over the past decade, advances in neural network architectures, coupled with the availability of large amounts of text data, have driven the development of increasingly complex and capable LLMs. Beginning with seminal works such as Google's BERT (Bidirectional Encoder Representations from Transformers) \citep{devlin2018bert} and OpenAI's GPT (Generative Pre-trained Transformer) \citep{radford2019language}, researchers have made substantial progress in enhancing LLMs' language understanding and generation capabilities \citep{liu2019roberta, raffel2020exploring}. These models leverage transformer architectures and self-attention mechanisms to capture long-range dependencies and contextual information, enabling them to generate coherent and contextually relevant text.

Further refinements, such as the introduction of GPT-3 \citep{brown2020language} with significantly larger parameter sizes and more sophisticated training regimes, have pushed the limits of LLM performance to unprecedented levels. These advancements have paved the way for LLMs to excel in a wide range of NLP tasks, including language translation \citep{costa2022no, kudugunta2024madlad}, text summarization \citep{lewis2019bart, zhang2020pegasus}, and question answering \citep{sanh2019distilbert, he2020deberta}.

\subsection{LLMs for mathematics}
The application of LLMs to mathematical reasoning tasks represents a recent and growing area of research within the machine learning community. Although LLMs have demonstrated remarkable proficiency in natural language understanding, their performance in mathematical reasoning tasks has historically lagged behind.

However, recent studies \citep{gou2023tora, shao2024deepseekmath} have shown promising results in using LLMs to solve mathematical problems. Transfer learning techniques, particularly fine-tuning pretrained LLMs on mathematical datasets, have emerged as effective strategies for enhancing LLMs' mathematical reasoning capabilities. By leveraging the knowledge encoded in pre-trained language models and adapting it to mathematical domains, researchers have achieved remarkable results on mathematical tasks.

Moreover, novel adaptation techniques, such as question paraphrase methods \citep{yu2023metamath} and tailored training objectives \citep{liu2023improving} aimed at enhancing LLMs' understanding of mathematical reasoning processes, have further advanced LLMs in mathematical applications. These techniques enable LLMs to effectively leverage existing data and develop robust reasoning abilities, thus expanding their utility in mathematical problem solving scenarios. However, these existing methods suffer from the problem of error propagation over long reasoning paths.

Overall, the application of LLMs in mathematics has immense potential to revolutionize the way mathematical problems are approached and solved. As research in this area continues to evolve, we can expect further advancements in LLMs' mathematical reasoning capabilities and their integration into diverse mathematical domains.

\section{Methodology}
The motivation for our methodology arises from the observation that while LLMs excel in various natural language processing tasks, their performance in mathematical reasoning remains suboptimal. This gap is primarily due to the inherent complexity of mathematical problems, which often requires intricate reasoning and logical deduction \citep{saxton2019analysing, lightman2023let}. Furthermore, the limited availability of annotated data in the mathematical domain \citep{liu2021roda, kumar2022practice} presents a significant obstacle to effectively training LLMs for such tasks.

To overcome these challenges, we propose a novel approach that combines question paraphrasing techniques with tailored training objectives to strengthen the mathematical reasoning capabilities of LLMs. An overview of our proposed pipeline is illustrated in Figure \ref{fig: pipeline}.

\input{figures/pipeline}

\subsection{Question Paraphrase}
Question paraphrasing is a crucial technique employed to augment existing data and enhance the model's ability to generalize across different linguistic forms of mathematical problems. This process involves leveraging the powerful generative capabilities of the GPT-4 model \citep{brown2020language} to produce diverse paraphrases for each question in the dataset. By generating multiple variations of the same question while preserving its semantic meaning, we aim to enrich the training data and expose the model to a wider range of linguistic structures and expressions commonly encountered in mathematical problem-solving scenarios.

\paragraph{Paraphrasing Questions} 
The GPT-4 model is utilized to generate the paraphrases for each question in the data set. Given a mathematical question, the model generates alternative phrasings that convey the same underlying mathematical concept. This step significantly enhances the diversity of the training data by presenting questions in various linguistic forms, such as synonyms, paraphrases, and syntactic variations.

To ensure the quality and semantic coherence of the paraphrased questions, each paraphrase, along with its corresponding answer, is inputted into the GPT-4 model. The model is then tasked with determining whether the paraphrased question-answer pairs match or not. This iterative validation process helps filter out the inconsistencies or semantic distortions, ensuring that only high-quality paraphrases are retained for training.

\subsection{Special Training Objectives}
Effective training objectives are critical for guiding a model's learning process and fostering a deeper understanding of mathematical concepts and reasoning. Our approach incorporates specialized training objectives designed to address the unique challenges of mathematical reasoning tasks.

During training, we start with a pretrained LLM and apply a supervised fine-tuning (SFT) objective on a carefully curated dataset of mathematical problems. This data set includes a diverse range of questions and solutions to ensure a comprehensive coverage of mathematical concepts and problem types. In the SFT phase, the model is trained to minimize the loss between its predicted answers and the correct solutions provided in the dataset, effectively aligning its outputs with the desired responses.

\paragraph{Rationale Re-Ranking (RR)}
The Rationale Re-Ranking (RR) objective aims to improve the model's ability to identify and reconstruct the correct reasoning path for solving mathematical problems. This objective involves shuffling the reasoning steps associated with a given problem and reordering the models to reconstruct the proper solution sequence. By training the model to recognize and organize the logical progression of mathematical solutions, the RR objective fosters a more structured, coherent, and accurate reasoning process.

\paragraph{Mistake Identification (MI)}
The MI objective focuses on improving the robustness and error tolerance of the model by training it to differentiate between correct and erroneous reasoning steps. During training, random modifications are introduced to induce errors in the reasoning process, such as changing numerical values or altering logical operators. The model is then trained to distinguish between correct and erroneous reasoning steps, thus learning to identify and rectify potential mistakes. This objective helps mitigate the risk of erroneous predictions and enhances the model's overall performance on mathematical reasoning tasks:

Integrating these training objectives into the fine-tuning process equips LLMs with enhanced capabilities for mathematical reasoning, thereby addressing the challenges posed by complex problem-solving scenarios.

\subsection{Training Process}
The training process of our proposed method consists of several stages designed to enhance the mathematical reasoning capabilities of LLMs. The process begins with data augmentation through question paraphrasing, where GPT-4 generates various paraphrases for each mathematical question. To ensure data set quality and integrity, we verify the consistency of the paraphrased questions and their corresponding answers, iterating this process to achieve high accuracy.

Next, we introduce specialized training objectives to improve the model's reasoning abilities. One such objective is Rationale Re-Ranking (RR), where reasoning steps are shuffled, and the model is tasked with predicting the correct sequence. This trains the model to understand and reconstruct logical progressions. Another objective is Mistake Identification (MI), where the model learns to distinguish between correct and intentionally altered reasoning steps, enhancing its ability to detect and correct errors in complex problem-solving scenarios. These objectives are incorporated within a multitask learning framework, with weighted losses assigned to each task to ensure balanced and effective training. This approach enables the model to optimize simultaneously for various aspects of mathematical reasoning, resulting in a more robust and versatile skill set.

By combining data augmentation, targeted training objectives, and a multitask learning framework, our method equips LLMs with strong mathematical reasoning skills, significantly improving their performance on challenging mathematical tasks.

\paragraph{Final training objective}
The final training objective is formulated as the weighted sum of $\mathcal{L}_{SFT}$, $\mathcal{L}_{RR}$, and $\mathcal{L}_{MI}$:
\begin{equation*}
\left.\begin{aligned}
\mathcal{L}_{final}(\theta) & = \lambda_{1}\mathcal{L}_{SFT} + \lambda_{2}\mathcal{L}_{RR} + \lambda_{3}\mathcal{L}_{MI}\\
\end{aligned}\right.
\end{equation*}

where $\lambda_{1}$, $\lambda_{2}$, and $\lambda_{3}$ are parameters that weight the importance of each loss, and $\mathcal{L}_{SFT}$, $\mathcal{L}_{RR}$, and $\mathcal{L}_{MI}$ are cross entropy losses.

\section{Experiments}
Mathematical reasoning challenges LLMs, requiring both natural language understanding and mathematical problem-solving skills. Evaluating the effectiveness of LLMs in mathematical reasoning is crucial for advancing state-of-the-art NLP techniques and expanding their applicability across diverse problem domains. This study seeks to assess the performance of various LLMs on mathematical reasoning tasks through targeted experiments.

\input{tables/main_results}

\subsection{Datasets}
We conducted experiments on four datasets specifically designed to assess LLMs' performance in mathematical reasoning.

\begin{itemize}
    \item \textbf{GSM8K} \citep{cobbe2021gsm8k}: A comprehensive dataset comprising mathematical problems covering a wide range of topics and difficulty levels.
    \item \textbf{MATH} \citep{hendrycksmath2021}: A curated collection of mathematical questions and solutions, designed to assess LLMs' ability to solve mathematical problems in various domains.
    \item \textbf{GSM8K-Hard} \citep{gao2022pal}: A subset of the GSM8K dataset containing challenging mathematical problems aimed at evaluating the robustness of LLMs under difficult scenarios.
    \item \textbf{SVAMP} \citep{patel-etal-2021-nlp}: A specialized dataset focusing on mathematical reasoning in the context of symbolic mathematics, presenting unique challenges for LLMs due to its symbolic nature.
\end{itemize}

\subsection{Base Models}
We employed four base models for our experiments, each representing a distinct architecture or variant of LLMs:

\begin{itemize}
    \item \textbf{LLama} \citep{touvron2023llama}: A baseline LLM model known for its strong performance in natural language understanding tasks.
    \item \textbf{LLama2} \citep{touvron2023llama2}: An enhanced version of LLama, incorporating improvements in model architecture and training methodology.
    \item \textbf{Mistral} \citep{jiang2023mistral}: A state-of-the-art LLM model specifically designed for mathematical reasoning tasks, leveraging advanced adaptation techniques.
    \item \textbf{Mixtral} \citep{jiang2024mixtral}: A high-capacity LLM model based on the Mixtral architecture, featuring eight times the parameter size of LLama for enhanced performance.
\end{itemize}

\paragraph{Training Details} We fine-tuned these base models on the aforementioned datasets with hyperparameters tailored to each model and dataset. The parameters are set as default to the ones used during pre-training. Note that we only experiment with the version of 7B parameters for each model. We adapt LoRA \citep{DBLP:conf/iclr/HuSWALWWC22} to make model fine-tuning more efficient. The rank and alpha are both set as 64. Additionally, we utilized specialized training objectives, such as Rationale Re-ranking (RR) and Mistake Identification (MI), to enhance the models' understanding of mathematical reasoning. We fine-tune each model with a single Nvidia A100 GPU and the runtime of each experiment is between 2 to 6 hours.

\subsection{Main Results}
Our experiments resulted in notable performance improvements across all four base models. Table \ref{tab: main_results} presents a summary of the main results. Our proposed special training objectives yielded an average performance boost of 4.25\% on GSM8K, 2.32\% on MATH, 6.21\% on GSM\_HARD, and 5.15\% on SVAMP datasets. Moreover, combining question paraphrase with these objectives further enhances the improvement to 7.32\% on GSM8K, 3.63\% on MATH, 7.72\% on GSM\_HARD, and 6.78\% on SVAMP. Notably, our methods have a more significant impact on relatively weaker models, likely because these models benefit more from structured reasoning guidance. This aligns with findings in previous LLM fine-tuning studies, where weaker models exhibit larger relative improvements when exposed to specialized training objectives.

These findings underscore the empirical effectiveness of our methodology in improving the reasoning efficiency and accuracy of LLMs. By assessing LLMs' performance in mathematical reasoning tasks, we contribute to the ongoing efforts to advance the state-of-the-art in natural language processing and pave the way for their application in diverse problem domains requiring mathematical reasoning abilities.

\section{Analysis and Discussion}
In this section, we delve into a comprehensive analysis and discussion of the experimental results, focusing on the effectiveness of our proposed method in enhancing LLMs' capabilities in mathematical reasoning tasks. We begin by dissecting the model's performance over varying numbers of reasoning steps, shedding light on the impact of our approach on problem-solving efficiency. Subsequently, we present findings from an ablation study aimed at elucidating the importance of individual components within our proposed methods. Finally, through a series of case studies, we illustrate both the successes and limitations of our approach, providing valuable insights for future research directions.

\subsection{Analysis on Reasoning Steps}

\input{figures/reasoning_steps}

We performed an in-depth analysis of model performance across varying reasoning steps to evaluate the effectiveness of our proposed method in solving complex mathematical problems. By examining performance at different depths of reasoning, we sought to highlight the impact of our approach on problem-solving efficiency. Figure \ref{fig: analysis_of_reasoning_steps} summarizes the model's performance on math questions requiring different reasoning steps. Overall, the results indicate that model performance declines as the number of reasoning steps needed to solve the problems increases. Although the improvement is minimal for questions requiring only a few reasoning steps (fewer than 4 steps), it becomes substantial for questions requiring longer reasoning chains (4–7 steps). Additionally, we note that the models struggle with questions that demand extremely long reasoning paths (more than 8 steps). 

Our analysis revealed a clear trend of improved model performance as the number of reasoning steps increased. This trend indicates that our proposed method effectively enhances the model's ability to solve complex mathematical problems, leading to more accurate solutions. Specifically, we observed substantial performance improvements on problems that involve multiple reasoning steps, which underscores the effectiveness of our approach in addressing complex problem-solving scenarios.

\subsection{Ablation Study}

\input{tables/ablation_study}

To further evaluate the importance of each component in our proposed methods, we performed an ablation study in which we systematically removed individual components and evaluated the model performance. 
Table \ref{tab: ablation_study} presents an ablation study of our proposed method on GSM8K and MATH datasets using Llama2 and Mistral as base models, respectively. Our results demonstrated that all components of our proposed methods are integral to improve model performance in mathematical reasoning tasks.

Specifically, when components such as question paraphrase techniques and specialized training objectives were integrated, we observed a marked increase in model performance, underscoring the importance of these components in facilitating effective mathematical reasoning. These findings emphasize the holistic nature of our proposed approach, wherein each component synergistically contributes to overall model performance.

\subsection{Case Study}

\input{figures/case_study}

We present a case study in Figure \ref{fig: case_study} to demonstrate the effectiveness of our method in solving complex mathematical problems. The case study includes a positive example showcasing successful problem-solving outcomes and a negative example highlighting challenges and limitations.

The positive example illustrates how our method empowers the model to navigate intricate mathematical problems and arrive at the accurate solution, even if it does not follow the exact reasoning path provided in the answer. This example validates the effectiveness of our approach in addressing real-world mathematical challenges.

In contrast, the negative example in Figure \ref{fig: case_study} reveals scenarios where our proposed method encounters limitations or fails to produce satisfactory results. Although the model follows the correct reasoning path, a common failure mode observed was arithmetic miscalculations despite correct reasoning paths. This suggests that while LLMs grasp mathematical structure, they struggle with precise computation—an issue that could be mitigated by integrating external calculation modules. This example identifies areas for improvement and prompts discussions on potential future research directions, including refining methodologies, leveraging external calculation tools, or exploring math verification approaches to enhance model performance in challenging scenarios.

\subsection{Discussion}

\paragraph{Data Efficiency and Generalization} Our question paraphrase strategy addresses a fundamental challenge in mathematical reasoning: the scarcity of diverse high-quality training data. By systematically transforming existing questions into linguistically varied forms while preserving their mathematical essence, we achieve improved generalization without requiring additional annotated examples. This approach is particularly valuable for specialized mathematical domains where expert annotation is costly and time-consuming. Our experimental results demonstrate that models trained with paraphrased questions exhibit enhanced robustness to linguistic variations, more closely mirroring the diverse ways in which mathematical problems may be encountered in real-world applications. This linguistic flexibility, combined with our specialized training objectives, enables models to focus on the underlying mathematical structures rather than become overly sensitive to specific phrasings. The data efficiency of our approach makes it particularly applicable to resource-constrained settings and suggests promising directions for self-supervised learning techniques that leverage mathematical invariance across different problem formulations.

\paragraph{Training-focused V.S. Test-time computational methods} In recent years, the Test-Time Computation (TTC) paradigm has gained significant traction as a means to improve the reasoning skills of LLMs, particularly for complex tasks. Although training-focused and TTC methods represent different strategies, they are not mutually exclusive. In fact, their synergistic integration holds significant promise for future advancements in mathematical reasoning for LLMs. The reasoning abilities that TTC techniques leverage are fundamentally learned during the training phase, encompassing both pre-training and fine-tuning. Training on high-quality mathematical data, including examples of step-by-step reasoning, provides the essential knowledge base that enables TTC methods (e.g., CoT \citep{wei2022chain} and Self-Refine \citep{madaan2023self}) to be effective. Furthermore, training processes can be optimized to produce models that are particularly adept at generating effective reasoning traces, which can then be further explored or verified during test time using TTC techniques. The quality of the underlying model, shaped by training, significantly influences how effectively it can utilize TTC methods.

\section{Conclusion}
In this study, we addressed the challenge of enhancing LLMs' capabilities in mathematical reasoning tasks. Our proposed method leverages innovative techniques to improve LLMs' understanding of mathematical concepts and reasoning processes. Through rigorous experimentation, we demonstrated the effectiveness of our approach in improving LLMs' performance on various mathematical problems.

Our research advances the field of natural language processing by providing a comprehensive approach to enhancing LLMs' capabilities in mathematical reasoning tasks. By identifying key factors influencing model performance and proposing effective solutions, we bridge the gap between natural language understanding and mathematical reasoning. Our study advances LLMs' reasoning capabilities, paving the way for broader applications in automated theorem proving, mathematical education, and AI-assisted research in scientific domains. Future work could explore integrating symbolic computation with LLMs to further enhance mathematical reliability.

\section*{Limitations}
While our study demonstrates promising results in improving the reasoning efficiency and accuracy of LLMs through data enhancement and fine-tuning, several limitations should be acknowledged. First, despite efforts to create a diverse data set through enhancement, inherent biases may persist in the training data. These biases could potentially skew the model's performance towards certain types of mathematical problems or reasoning patterns. Second, although our fine-tuned model improves test performance, its generalization to out-of-distribution mathematical problems remains uncertain. Future work could explore curriculum learning strategies or hybrid neural-symbolic approaches to mitigate this limitation. Further investigation is needed to assess the model's capabilities in tackling advanced mathematical concepts or interdisciplinary problems that deviate significantly from the training examples.

\bibliography{acl_latex}




\end{document}

%% file: figures/illustration_example.tex
\begin{figure*}[ht]
\centering
\includegraphics[width=0.7\linewidth]{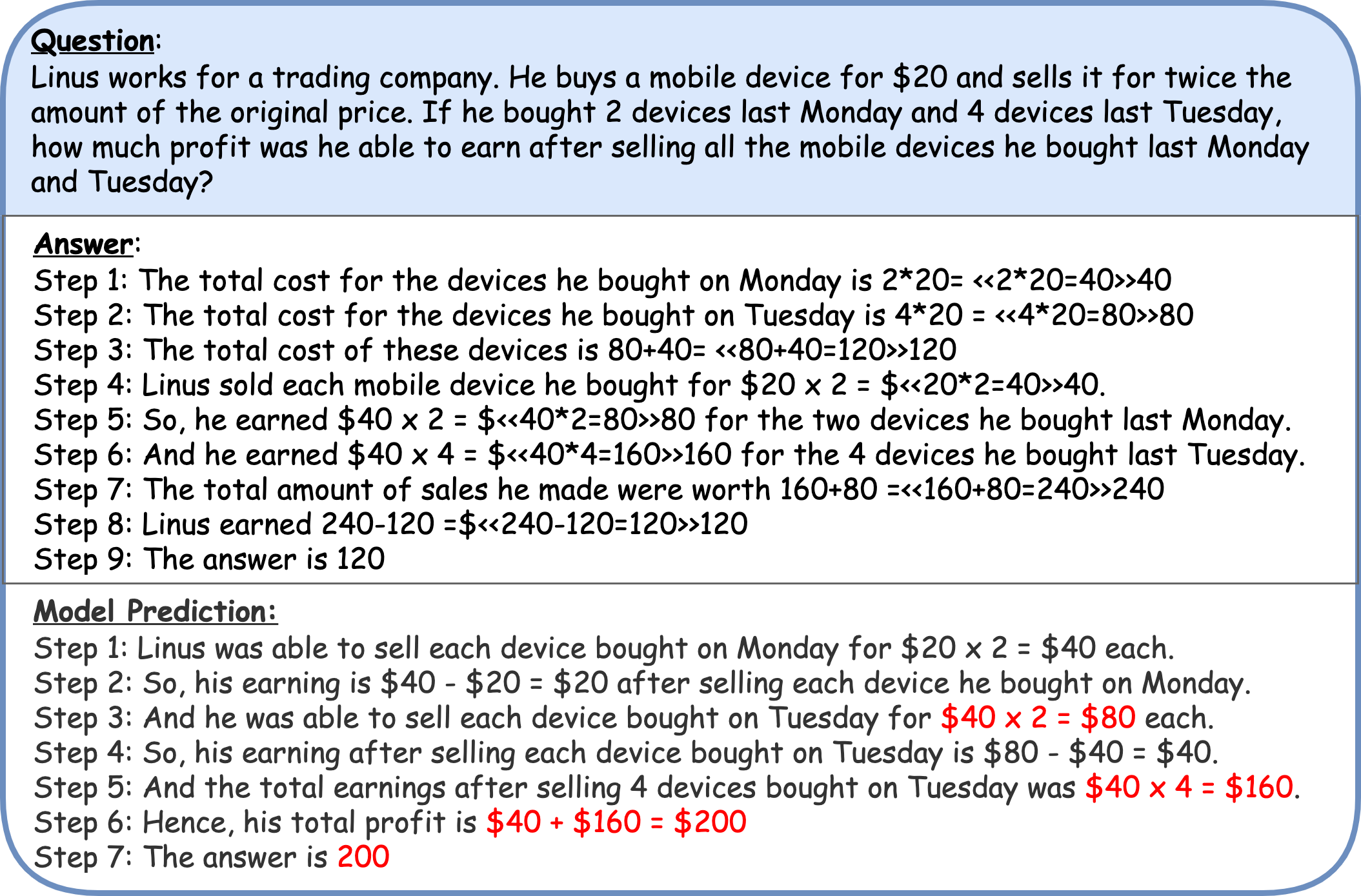}
\caption{Llama2-7B model prediction on an example from the GSM8K dataset.}
\label{fig: illustration_example}
\end{figure*}

%% file: figures/pipeline.tex
\begin{figure*}[ht]
\centering
\includegraphics[width=0.9\linewidth]{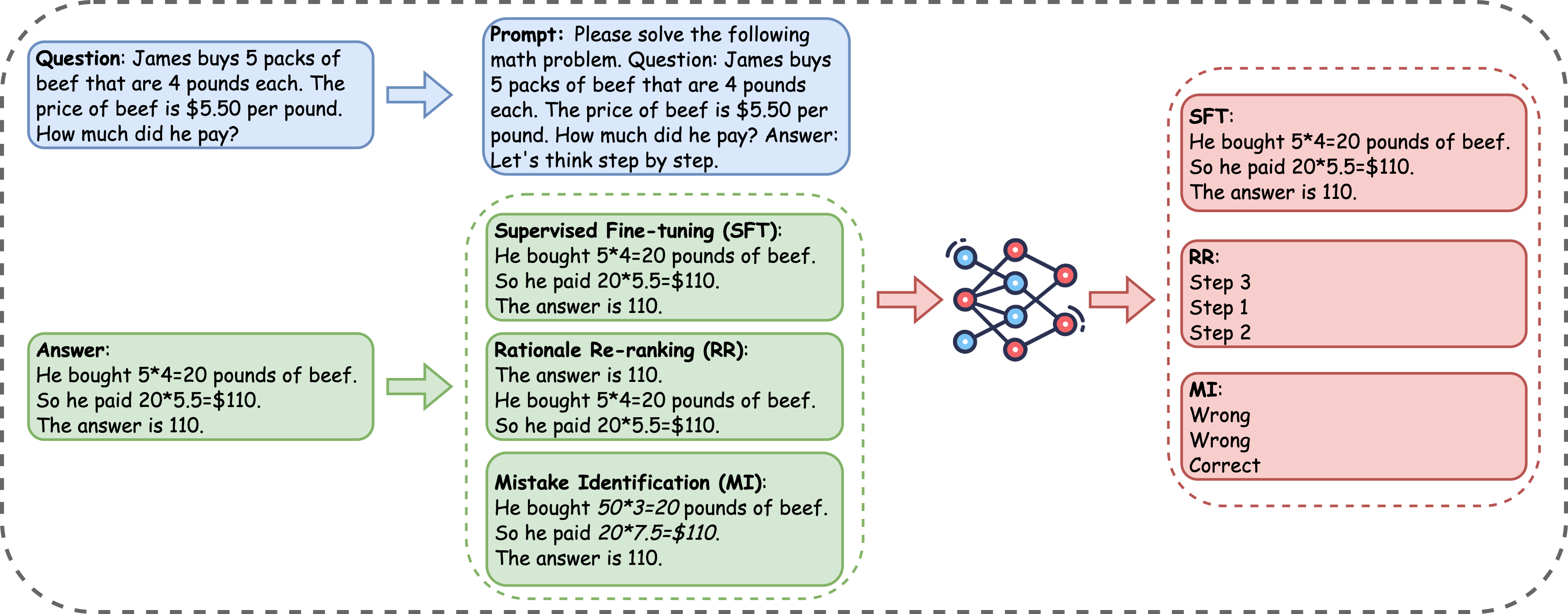}
\caption{The overview of our proposed pipeline.}
\label{fig: pipeline}
\end{figure*}

%% file: tables/main_results.tex
\begin{table*}
    \centering
    \resizebox{0.75\linewidth}{!}{
        \begin{tabular}{lcccc}
            \toprule
            \multirow{2}{*}{\bf Method} & \multicolumn{4}{c}{\bf Dataset} \\
            \cmidrule(lr){2-5}
                & \bf GSM8K & \bf MATH  & \bf GSM\_HARD & \bf SVAMP \\
            \midrule
            \multicolumn{5}{l}{\textbf{\textit{Baselines (SFT Only)}}} \\
            \midrule
            Llama-7B \citep{touvron2023llama}        & 32.07 & 5.60 & 23.43 & 38.24 \\
            Llama2-7B \citep{touvron2023llama2}      & 36.92 & 5.68 & 26.72 & 41.07 \\
            Mistral-7B \citep{jiang2023mistral}     & 58.68 & 14.08 & 55.42 & 50.25 \\
            Mixtral-8x7B \citep{jiang2024mixtral}   & 65.44 & 30.13 & 64.93 & 68.73 \\
            \midrule
            \multicolumn{5}{l}{\textbf{\textit{Proposed Method (SFT + MI + RR)}}} \\
            \midrule
            Llama-7B \citep{touvron2023llama}       & 37.15 & 6.91 & 29.71 & 44.22 \\
            Llama2-7B \citep{touvron2023llama2}      & 43.04 & 7.65 & 38.74 & 49.03 \\
            Mistral-7B \citep{jiang2023mistral}     & 62.87 & 15.98 & 60.02 & 56.87 \\
            Mixtral-8x7B \citep{jiang2024mixtral}   & 67.04 & 34.22 & 66.86 & 68.76 \\
            \midrule
            \multicolumn{5}{l}{\textbf{\textit{Proposed Method + Question Paraphrase}}} \\
            \midrule
            Llama-7B \citep{touvron2023llama}       & 41.74 & 7.64 & 37.87 & 46.14 \\
            Llama2-7B \citep{touvron2023llama2}      & 46.86 & 9.05 & 32.96 & 53.04 \\
            Mistral-7B \citep{jiang2023mistral}     & 65.82 & 17.27 & 63.25 & 58.03 \\
            Mixtral-8x7B \citep{jiang2024mixtral}   & 67.96 & 36.07 & 67.28 & 68.19 \\
            \bottomrule
        \end{tabular}
    }
    \caption{Experimental results of base models on different datasets, respectively. Scores are calculated with the accuracy metric. \\}
    \label{tab: main_results}
\end{table*}

%% file: figures/reasoning_steps.tex
\begin{figure*}[ht]
\centering
\subfigure[Llama2-7B on different reasoning steps]{
    \begin{minipage}[t]{0.45\linewidth}
    \centering
    \includegraphics[width=1\linewidth]{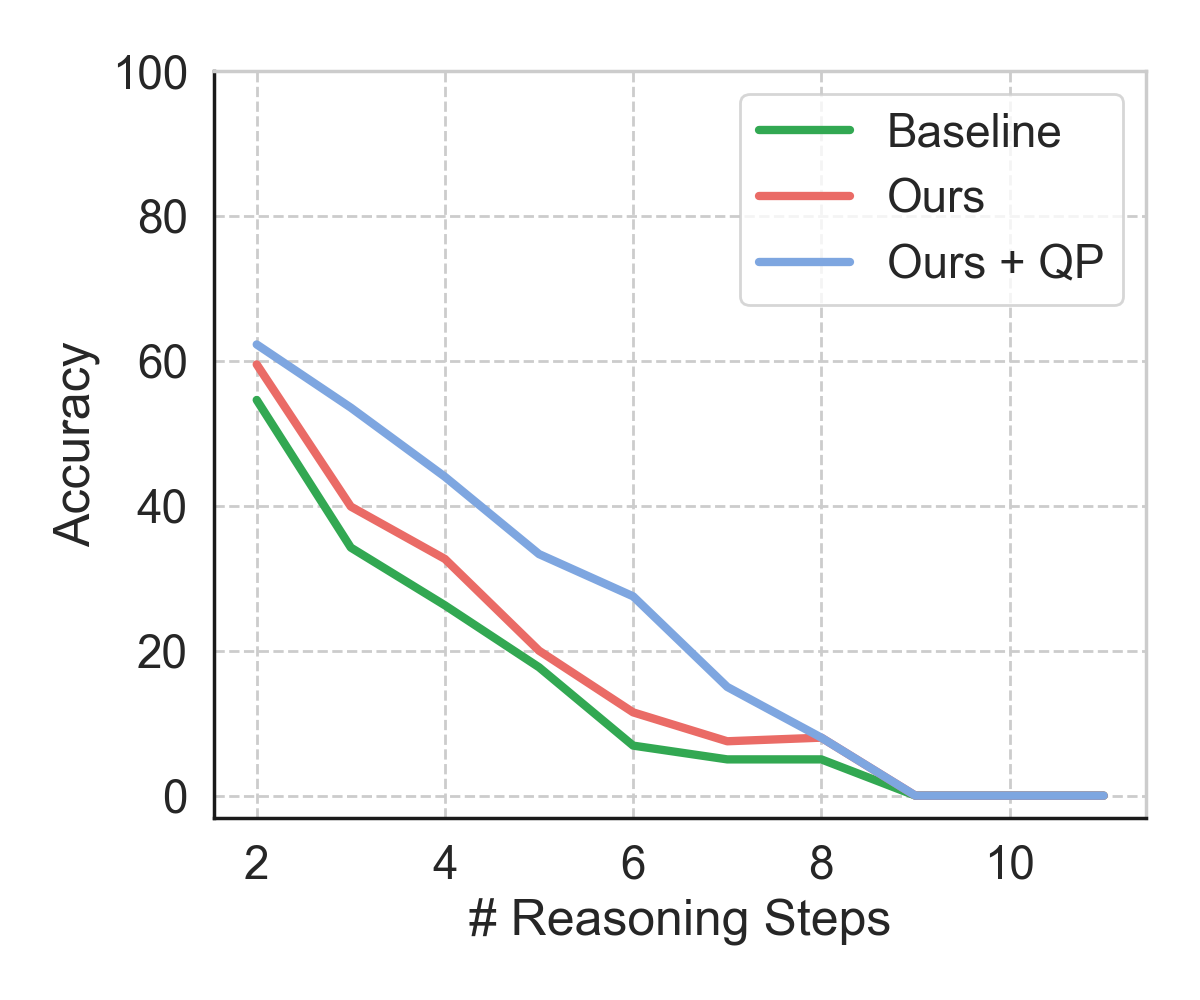}
    \end{minipage}
    \label{fig: llama2_reasoning_steps}
}
\subfigure[Mistral-7B on different reasoning steps]{
    \begin{minipage}[t]{0.45\linewidth}
    \centering
    \includegraphics[width=1\linewidth]{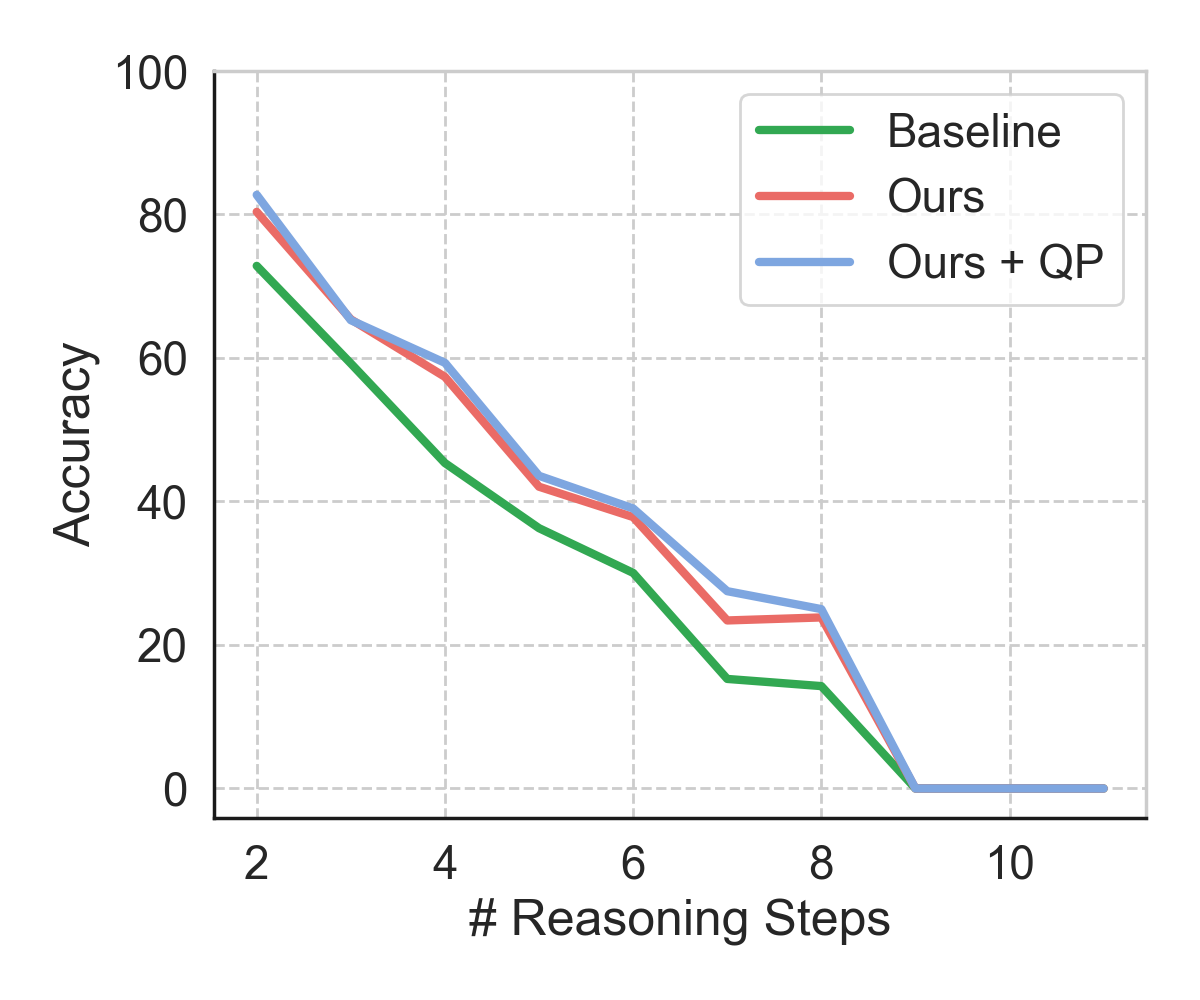}
    \end{minipage}
    \label{fig: mistral_reasoning_steps}
}
\caption{Analysis of Llama2 and Mistral on different reasoning steps, respectively. X-axis is the number of reasoning steps required to solve a math question and Y-axis the accuracy score. We use supervised fine-tuning as the baseline, and compare it with our proposed method using special training objective. QP stands for question paraphrase.}
\label{fig: analysis_of_reasoning_steps}
\end{figure*}

%% file: tables/ablation_study.tex
\begin{table}
    \centering
    \resizebox{0.97\linewidth}{!}{
        \begin{tabular}{lcccc}
            \toprule
            \multirow{2}{*}{\bf Method} & \multicolumn{2}{c}{\bf GSM8K} & \multicolumn{2}{c}{\bf MATH}\\
            \cmidrule(lr){2-3} \cmidrule(lr){4-5}
                & \bf Llama2-7B    & \bf Mistral-7B  & \bf Llama2-7B & \bf Mistral-7B \\
            \midrule
            Baseline (SFT) & 36.92 & 58.68 & 5.68 & 14.08 \\
            \midrule
            \hspace{0.15 cm} + RR & 38.94 & 59.06 & 5.83 & 13.07 \\
            \hspace{0.15 cm} + MI & 40.01 & 61.57 & 7.83 & 15.24 \\
            \hspace{0.15 cm} + MI + RR & 43.04 & 62.87 & 7.65 & 15.98 \\
            \hspace{0.15 cm} + MI + RR + QP & 46.86 & 65.82 & 9.05 & 17.27 \\
            \bottomrule
        \end{tabular}
    }
    \caption{Ablation Study of our proposed method on GSM8K and MATH using Llama2-7B and Mistral-7B as base models, respectively. SFT, RR, MI, and QP stand for supervised fine-tuning, rationale re-ranking, mistake identification, and question paraphrase, respectively. Scores are calculated with the accuracy metric.\\}
    \label{tab: ablation_study}
\end{table}

%% file: figures/case_study.tex
\begin{figure*}[ht]
\centering
\subfigure{
    \begin{minipage}[t]{0.48\linewidth}
    \centering
    \includegraphics[width=1\linewidth]{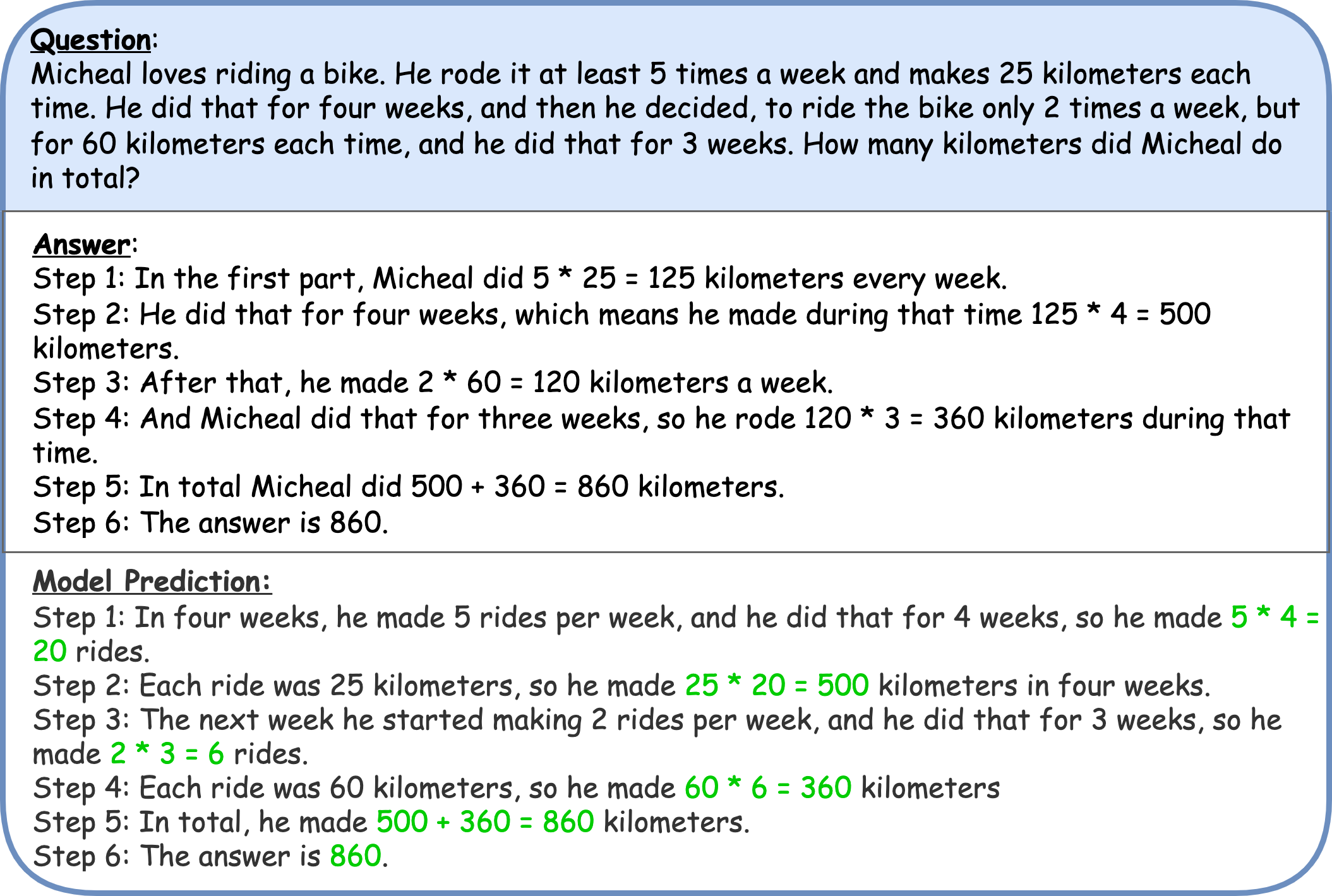}
    \end{minipage}
    \label{fig: case_study_pos}
}
\subfigure{
    \begin{minipage}[t]{0.48\linewidth}
    \centering
    \includegraphics[width=1\linewidth]{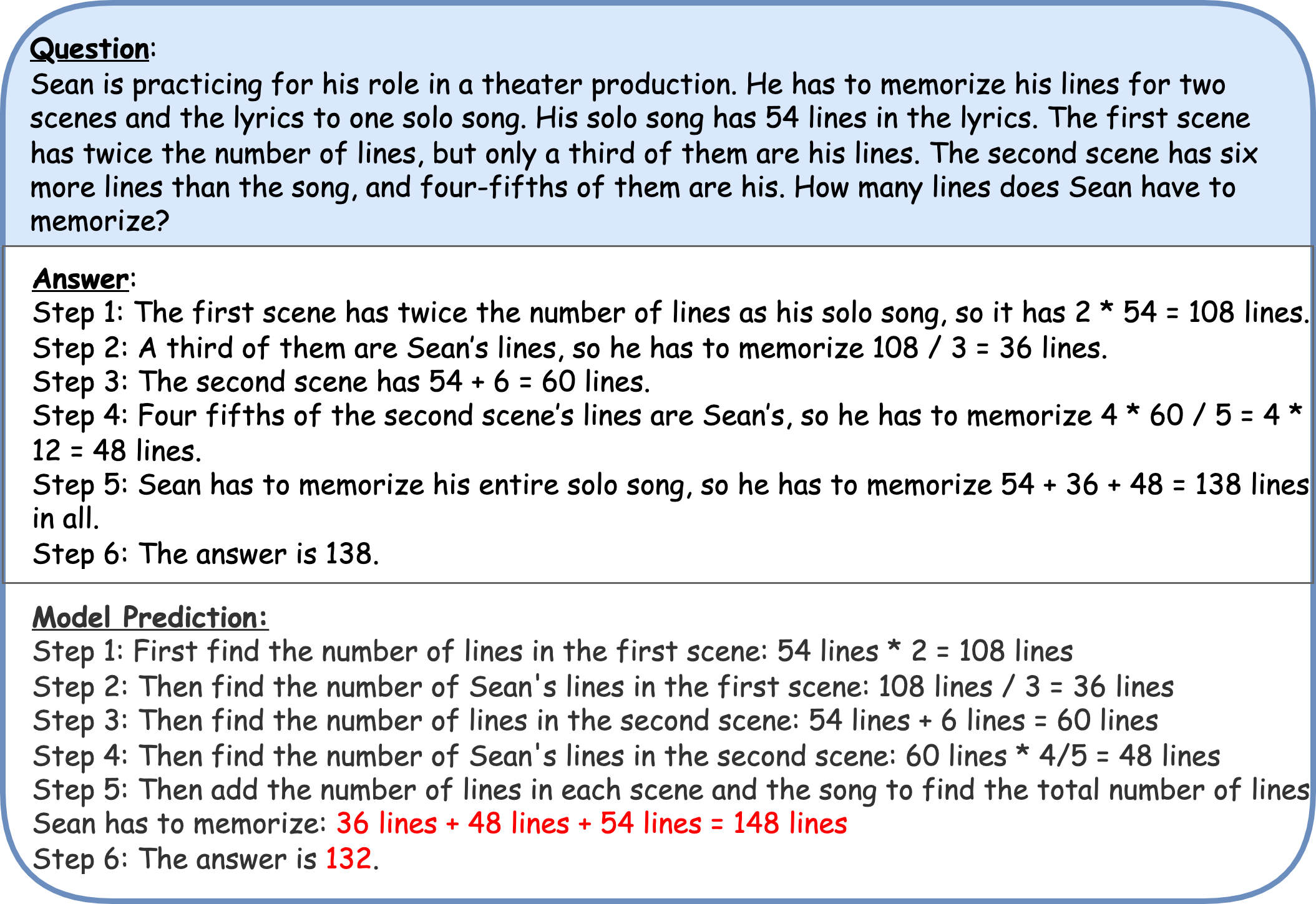}
    \end{minipage}
    \label{fig: case_study_neg}
}
\caption{Case study on the GSM8K dataset using the Mistral-7B Model. The example on the left is positive while the example on the right is negative.}
\label{fig: case_study}
\end{figure*}